# Automated Question Answer medical model based on Deep Learning Technology

Abdelrahman Abdallah
MSc Machine Learning & Data Science, Satbayev University
Almaty, kazakhstan
abdoelsayed2016@gmail.com

Mahmoud Kasem
Teaching Assistant, Information Technology, Assiut University
Assiut, Egypt
mahmoud.salah@aun.edu.eg

Mohamed Hamada
Associate professor, IS department
International IT University
Almaty, Kazakhstan
Dr.mmhamada@gmail.com

Shaymaa Sdeek
dept. Information System Assiut University
Qena, Egypt
shaymaaragab123@gmail.com

## ABSTRACT
Artificial intelligence can now provide more solutions for different problems, especially in the medical field. One of those problems the lack of answers to any given medical/health-related question. The Internet is full of forums that allow people to ask some specific questions and get great answers for them. Nevertheless, browsing these questions in order to locate one similar to your own, also finding a satisfactory answer is difficult and time-consuming task. This research will introduce a solution to this problem by automating the process of generating qualified answers to these questions and creating a kind of digital doctor. Furthermore, this research will train an end-to-end model using the framework of RNN and the encoder decoder to generate sensible and useful answers to a small set of medical/health issues. The proposed model was trained and evaluated using data from various online services, such as WebMD, HealthTap, eHealthForums, and iCliniq.

## 1. INTRODUCTION
The Question Answering System[10] is an intelligent information system that retrieves correct answers to natural language questions over a short period of time. Such systems are especially useful in the field of bio-medicine where medical literature continues to develop.

Issue answering sites are both gaining attention to patients and doctors. Patients receive immediate support and trustworthy responses for both basic and complex health issues. Physicians will improve their popularity among their colleagues , strengthen their clinical expertise from experiences with other renowned physicians, as well as draw more new patients.There is a language divide between knowledge seekers and health care professionals to close this divide, using local mining and global approaches to learning.A large number of medical documents have been stored in their libraries and, in most cases, will the user find useful answers directly by searching these record archives instead of waiting for the experts? Answer or search through a list of potentially important Web documents.

Question Answering (QA) programs are very useful, as most of the problems associated with deep learning can be modeled as a problem answering questions.Consequently, in today's computer science, the field is one of the most researched areas.The last few years have seen major advances and advancements in the state of the art, much of which can be traced to the emerging Deep Learning or specifically Deep Neural Networks, in many topics such as object detection and image classification using Convolutional Neural Networks (CNN)[4, 5], understanding speech using Recurrent Neural Networks(RNN)[1,2], and named entity recognition also using RNN [3].

Most medical NLP studies have attempted to apply general language models to medical tasks. These models are not trained on medical information; however, they make mistakes that reflect this. In this study, we use the scope of information stored inside a medical question-answer pair to insert medical knowledge into the model to increase the features in these general language models. Question-answer tasks such as Machine Translation have been found  Great successful in using Deep Neural Networks is more difficult to achieve, but recent research has shown very promising results using RNN.Specifically, the end-to-end approach known as Neural Machine Translation (NMT)[8, 9], which uses the generic encoder decoder framework proposed in [7], encoding sequences for fixed-sized latent vector representations from which a decoder produces a new sequence, has been shown to beat state-of-the-art phrase-based translation systems. Another difficult issue related to the NMT issue, which Deep Learning has not yet solved, is the so-called "Text Generating Neural Network" (TGNN) issue.this problem considers the generation of a suitable human-like answer/response to a single given statement, post, or question.

In this project, the TGNN problem is considered limited to the topics of health and medical question answers. the problem of generating answers to given medical questions was studied.

The problem considered in this project is the lack of available answers to any given medical/health-related question. The Internet is full of forums allowing people to ask these questions, from which some of them retrieve great answers. However, browsing these questions to locate one similar to your own, and also finding one with a satisfying answer is a difficult and time-consuming task, and posting the question yourself is even worse. By automating the process of generating answers to these questions, and thereby creating a kind of digital doctor, one could solve this problem. The problem thereby limits to the TGNN problem, i.e. given a medical/health-related question, a single suitable answer should be generated.

The TGNN problem and the lack of available Q/A data within medical/health topics compared to data sources for general conversations such as Twitter and Weibo, creating an end-to-end well-performing digital doctor is obviously very optimistic. However, since the amount of available general medical/health text from Wikipedia etc. is very large, hence generating somewhat useful answers for a limited set of questions seems possible. Thus, the goal of this project is to train an end-to-end model using RNN's and the encoder-decoder framework, which should be able to generate sensible, somewhat useful answers to a small set of given medical/health questions.

We used the MedQuAD collection of 47,457 question-answer pairs from trusted medical sources that introduce and share in the A Question-Entailment Approach To Question Answering[15] . Also we collected datasets from online services like WebMD, HealthTap, eHealthForums, and iCliniq.

The following section defines the related work on question answering, question similarity, and entailment. Section 3 presents end-to-end models using RNN's and the encoder-decoder framework, which should be able to generate sensible, somewhat useful answers to a small set of given medical/health questions. Section 4 provides experimental results on the test data obtained on LiveQA medical questions and concludes remarks are given in Section 5.

## 2. RELATED WORK

(Asma Ben Abacha , 2019 et al.)[15] In this paper, propose a novel QA approach based on Recognizing Question Entailment (RQE) and describes the QA system and resources that they built and evaluated on real medical questions. First, they compared machine learning and deep learning methods for RQE using different kinds of datasets, including textual inference, question similarity and entailment in both the open and clinical domains. Second, they combine IR models with the best RQE method to select entailed questions and rank the retrieved answers.

(Andrea Andrenucci , 2008 )[16] This paper has discussed three main techniques within QA and has pointed out approaches that are more suitable for medical applications: the deep NLP approach and the template-based approach. The template-based approach is the most viable commercially and fits Web-based medical applications that are aimed at retrieving multilingual content in different multimedia formats. Its high recall level makes it a technique that fits users who are more interested in retrieving complete sets of answers rather than a few very precise answers.

The deep NLP approach provides a dialogue that better resembles human-to-human conversation and also delivers more reliable answers. It fits areas where the precision of the retrieved information is crucial, e.g. in decision-support or evidence-based medicine. Furthermore, IR enhanced by shallow NLP is more appropriate as a search tool for larger or open domains as the Web because it exploits data redundancy. However, it can mainly retrieve factual answers unlike the NLP and template-based approaches, which support more complex types of questions such as requests for advice giving.

(Junqing He , 2019)[17] They implemented a large-scale Chinese medical QA dataset in this paper, and cast the challenge into a question of semantic matching. We compare different CWS devices and input modules, too. Match Pyramid performs better among the two state-of-the-art deep matching neural networks. The results also show the efficacy of the proposed clustered semantic representation module. Automatic answering of medical questions is a special form of answering questions (QA) that requires scientific or clinical information. Due to inadequate practitioners and difficult access to hospitals for certain people, there is a growing need to implement advanced automated medical QA systems. According to an American health survey, the general search engines were used by 59 percent of U.S. people on the Internet for health information, 77 percent of whom. However, to find the desired information, they must filter numerous results of their queries. Health consultation websites have arisen for this reason, with thousands of medical practitioners and active patients answering user-suggested questions. But this kind of service fails to provide customers with immediate and correct responses, which for certain patients is intolerable. In addition , medical QA systems also benefit physicists by providing fellows with previous answers as a reference.

(Xiao Zhang , 2018 )[18] In this work,they implement a question-answering activity called MedQA to study answering questions using expertise in a large-scale set of documents in clinical medicine. MedQA 's aim is to respond to real-world questions with a broad reading understanding. Say their SeaReader solution — a modular end-to-end read comprehension platform based on LSTM networks and a dual-path architecture. The novel dual-path attention models flow information from two perspectives and have the ability to read individual documents simultaneously and integrate information across multiple documents. SeaReader obtained a substantial improvement in accuracy on MedQA over competing models in experiments. Additionally, a set of novel techniques are developed to illustrate the understanding of the SeaReader question-answering process.

## 3. PROPOSAL WORK

The goal of TGNN is to generate answers to a question by a Recurrent Neural Network (RNN) is a network that operates on a sequence and uses its own output as input for subsequent steps. A seq2seq network is a model that consists of two RNNs called the encoder and decoder. The input sequence is read by the encoder and a single vector is output, and the vector is read by the decoder to generate an output sequence.

### 3.1 Data

For this project, the following two text datasets were used:

First, Medical Question Answering Dataset (MedQuAD): MedQuAD contains 47,457 pairs of medical question-answer, generated from 12 NIH websites (e.g, cancer.gov, niddk.nih.gov, GARD and MedlinePlus Health Topics). The list includes 37 types of questions (e.g, treatment, diagnosis, side effects) related to illnesses, medications, and other medical institutions such as examinations.

Second, we collected datasets from websites like WebMD, HealthTap, eHealthForums, and iCliniq: The question/answer (Q/A) pairs are scraped from WebMD, HealthTap, eHealthForums, and iCliniq websites.Contains classified user-specific questions and their related expert responses.

Examples of a Q/A pair are as follows:

1) **How does support stocking help impre low blood pressure symptoms during exercise?**

    Support hose improves the return of blood from the leg veins to the central circulation and can help improve blood pressure. There is also a possible mild improvement in blood pressure from mild general vascular compression in the lower extremities.

2) **acid reflux shortness of breath symptoms.**

    TAcid reflux: symptoms can include shortness of breath, chest discomfort, and burping. These symptoms are more similar to those experienced due to heart disease (angina, heart attack) than to symptoms of COPD. There may be some overlap with symptoms from all three conditions. If a person is unsure about what may cause these types of symptoms, there should be a medical workup to determine the origin.

### 3.2 Models

We trained four types of models:

- GRU/LSTM: Simple sequence-to-sequence model based on GRU/LSTM Recurrent Network, respectively.
- Reverse: Apply Bi-directional LSTM to the encoder part
- Embeddings: Apply Fasttext[18] word embeddings (300D)
- Attention: Apply Attention Mechanisms to the decoder part

The general Encoder-Decoder is used for each model, but each with a different architecture in either the encoder or the decoder shown in Fig .1. The seq2seq network encoder is an RNN outputting some value from the input sentence for each word. The encoder outputs a (context) vector and hidden condition for each input word. The encoder's last hidden state would be an initial hidden decoder condition. The decoder is another RNN that takes the vector(s) of the encoder output, and produces a sequence of words to create the translation.

Sometimes this last output is called the context vector, as it encodes context from the whole sequence. This context vector is utilized as the decoder's initial hidden state. The decoder is given an input token and hidden condition at each decoding stage. The initial input token is the start of the string token, and the first hidden condition is the context vector (= the last hidden state of the encoder). The decoder continues to generate words until a token is output, which represents the end of the sentence.

We will also use an Attention Mechanism in our decoder to help it take care of certain parts of the input when generating the output.

The sequence-to-sequence network training method is distinct from the inferencing method. The difference between training and inference is due in essence to the self-regressive property.

Auto-regressive: infers (or predicts) the present value by referring to its own values in the past. As shown below, the current time-step output value $y\_t$ is determined by both the encoder's input sentence (or sequence) X and the previous time-step output ($y\_1,...y\_t-1$).

$$y_i = \text{argmax}_y P(y|X, y < i); \text{ where } y_0 = \text{SOS Token} \quad (1)$$

In the past, if our decoder makes a false prediction, it can lead to a greater false prediction.Teacher Forcing: Thus, we practice using a method called Teacher Forcing, which is the idea of using the actual target outputs as the next input, rather than using the guess of the decoder as the next input.

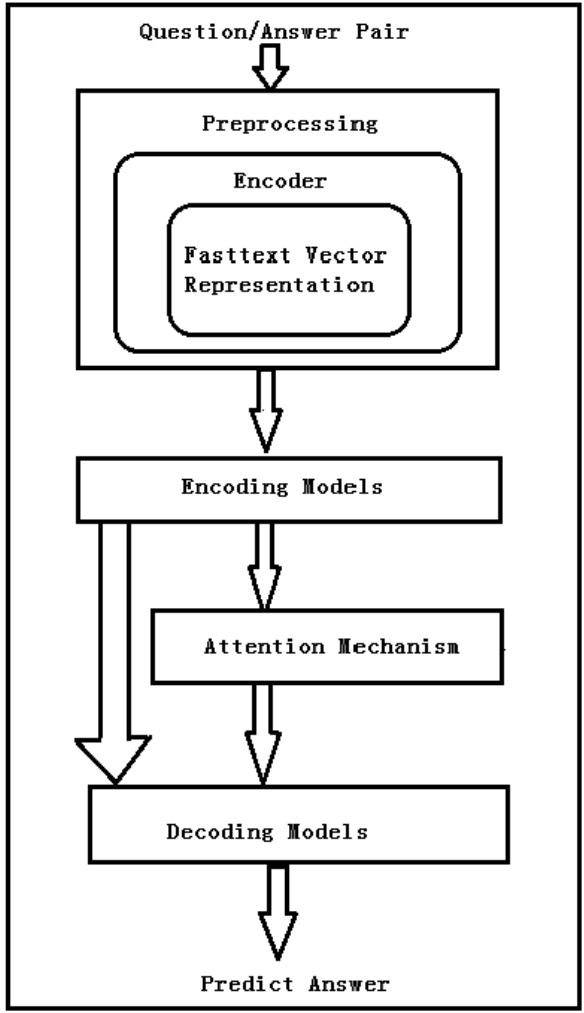

Fig. 1: Encoder and Decoder Architecture

In the training mode, the network seq2seq considers that we already know all the answers (, the last time-step outputs). Through inference mode, the network seq2seq takes the input from the last time-step output.

$$\hat{y} = argmax_y P(y|X, y < t; \theta); where \quad (2)$$
$$X = x_1, ...., x_n \text{ and } Y = y_1, ...., y_n$$

If only the context vector between the encoder and also the decoder is passed, that single vector will carry the burden of encoding the full sentence. Attention allows the network of decoders to "focus" on another part of the encoder output for every step of the decoder's own outputs.

First, we calculate a series of weights for the focus. These are multiplied by vectors of the encoder output to construct a weighted combination. The result will contain information about that particular part of the input sequence, thereby helping the decoder pick the appropriate output terms.

A bach matrix-matrix product of matrices stored in the output of the decoder and output of the encoder are used to measure the attention weights. Because the training data includes sentences of all sizes, to actually build and train this layer, we need to select the maximum sentence length (input length, for encoder outputs) that it will refer to. Maximum length sentences will use all the weights of focus, while shorter sentences use only the first few.

The training of the networks was performed using mini-batches of size 64 with the Stochastic Gradient Descent algorithm. The data were split into training (70%) and test (30%), and the networks were all regularized using early stopping and with gradient norms, and for some of the networks the dropout mechanism and weight decay. The network parameters were all initialized at random from a uniform distribution in the interval $[-\sqrt{3}/3, +\sqrt{3}/3]$ with $d$ is the dimensionality of the input at a given layer, which has been shown to be a good initialization strategy for deep networks [20].

A general problem with training RNN's by feeding the true token to the network at each time-step is the discrepancy between training and inference/prediction. During training, the network always gets the correct token as input at a given time-step $t$ independently of what the predicted token in time step $t-1$ was hence, it's mistakes do not propagate over time. During inference, the true token is not available; hence, the network always gets the previous predicted token as input. I.e. When the networks make a mistake by predicting an incorrect token, the mistake propagates throughout the time steps, and the predicted sequence diverges completely from the correct; Especially if the incorrect prediction is in an early time-step.

This problem is stated in [21] and an alternative training strategy claimed to solve the problem, refered to as Scheduled Sampling, is proposed. Scheduled Sampling is a training strategy for the RNN's proposed in [21], which uses a mix between the correct token and the previously predicted token as input at a given time-step $t$ during training. Given the $i^{th}$ mini-batch out of an estimated $i_{max}$ number of mini-batches required for convergence, then for each time-step t during training, denotes ε the probability of using the correct token as input with $1- ε_i$ being the probability of using the infered token from the previous time-step.

### 3.3 Experiments

The proposed and tested models have all been implemented using the Pytorch library [22] for Python, which allows for the transparent use of highly optimized mathematical operations on GPU's through Python.PyTorch is a Python package that provides two high-level features: tensor computation (like NumPy) with strong GPU acceleration and deep neural networks built on a tape-based autograd system. The experiments were run on a machine with 2x "Intel(R) Xeon(R) E-5-2680" CPU's and 4x "NVIDIA Tesla k20x"

### 3.4 Result

As discussed above, we have four models, each model is trained in different length's of question/answer pairs with 50, 75, 100 characters. We compared the losses between the different models in the training shown in Fig. 2, Fig. 3, Fig. 4, Fig. 5, and Fig. 6.

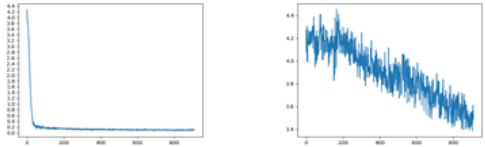

(a) Sequence length of 50.   (b) Sequence length of 75.

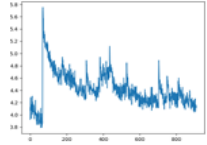

(c) Sequence length of 100.

Fig. 2 GRU losses for sequences length 50, 75, and 100.

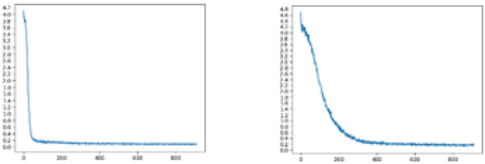

(a) Sequence length of 50.   (b) Sequence length of 75.

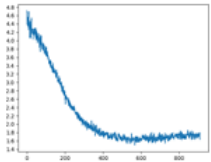

(c) Sequence length of 100.

Fig. 3 LSTM losses for sequences length 50, 75, and 100.

For compare the performance of the different models, the BLEU[19] score , it is a popular metric within machine translation for evaluating the similarity between two strings. hence it is used to evaluate the quality of a generated translation vs. a true translation in Machine Translation or Sequence to Sequence tasks , based on n-gram precisions between the two. In order to penalize trivial candidates containing e.g.

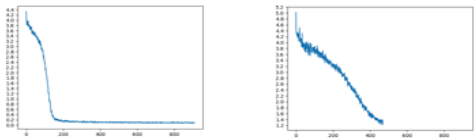

a) Sequence length of 50.   (b) Sequence length of 75.

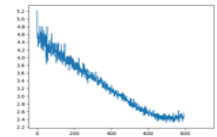

(c) Sequence length of 100.

Fig. 4 Embeddings with Bi-directional LSTM for sequences length 50, 75, and 100.

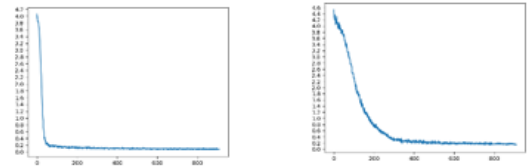

a) Sequence length of 50.   (b) Sequence length of 75.

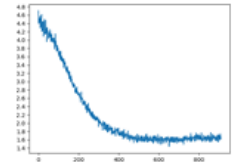

(c) Sequence length of 100.

Fig. 5: Bi-directional LSTM for sequences length 50, 75, 100.

only the same word, which will get a high n-gram precision, a "modified n-gram precision" is used. The modified n-gram precision is defined as follows: Given a candiate sentence C and a set of reference sentences

$$R = R1, R2, ...$$

Then for each n-gram in C count the maximum occurrence in each reference R in R and the number of occurrences in the candidate C. These two counts are summed up for all n-grams in C, and the modified precision is then given as the total number of maximum relevant n-gram occurrences in the references R divided by the total number of n-gram occurrences in the candidate.

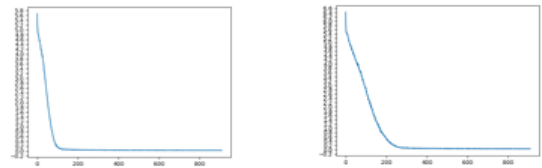

a) Sequence length of 50.   (b) Sequence length of 75.

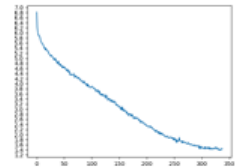

(c) Sequence length of 100.

Fig. 6: Attention mechanisms for sequences length 50, 75, 100.

We evaluated model performance by the accuracy of choosing the best candidate. The results are shown in Table 1. and Table 2. Our models clearly outperform baseline models by a large margin. Tables show the BLEU in the training data and testing data.

Table 1. BLEU for Training data

|  | 50 Chars | 75 Chars | 100 Chars |
|---|---|---|---|
| GRU | 84.61 | 75.55 | 65.53 |
| LSTM | 83.54 | 74.17 | 64.33 |
| Bi-LSTM | 84.96 | 79.12 | 69.29 |
| Embeddings with Bi-LSTM | 85.12 | 80.71 | 72.90 |
| Attention | 86.51 | 81.96 | 73.11 |

Table 2. BLEU for Testing data

|  | 50 Chars | 75 Chars | 100 Chars |
|---|---|---|---|
| GRU | 80.25 | 70.74 | 52.65 |
| LSTM | 80.67 | 69.55 | 51.71 |
| Bi-LSTM | 81.31 | 72.55 | 53.96 |
| Embeddings with Bi-LSTM | 82.23 | 77.96 | 54.06 |
| Attention | 83.11 | 78.80 | 59.31 |

Below list shows examples from test data. First example from the Attention model with sequence length of 75. Second example from Attention model with sequence length of 50. Third and fourth examples from Attention model with sequence length of 100.

1) **what are reasons for possible abnormal results of a pericardiocentesis ?**
   Ground Truth Text: common diagnoses are renal failure heart failure infection and cancer .
   Predicted text: common diagnoses are renal failure heart failure infection and cancer .

2) **is it possible for men to get breast cancer ?**
   Ground Truth Text: about breast cancer occurrence in men .
   Predicted Answer: men get breast cancer as often as women .

3) **what does lae mean if ecg is adnormal rbbb ?**
   Ground Truth Text: left atrial enlargement is independent of the right bundle branch block .
   Predicted Answer: left atrial enlargement is independent of the right bundle branch block .

4) **may i perform a blood donation if i m taking bisoprolol for arrhythmias ?**
   Ground Truth Text: Usually, they don't take blood from people taking vasoactive drugs .
   Predicted Answer: But if you have lightheadedness you should not .

## 4. Conclusion

From the analysis of this project, it can be inferred that training an end-to-end network, using the encoder-decoder system similar to the one used for TGNN, is a very difficult task to generate answers to medical questions, though somewhat possible. A bidirectional recurrent neural network with a final BLEU score of 86.51, 81.96 and 73.11 has been shown to be possible. To learn some set template answers to specific types of questions, and also to learn the answers to specific types of questions that need more user knowledge. In addition , a large fraction of the generated answers are spelled correctly and understandable answers, so the model has gained some understanding of syntax and semantics.

The early epochs showed some slight improvements in convergence speed; however, each epoch was significantly slower due to the extra data and the decoder handling the "mode-switching" between the language model-mode and the answer generating mode.

It has also been found that the BLEU score is not a suitable performance evaluation metric for the TGNN problem, since the set of possible valid answers to a given question is much larger and more complex than that one for bilingual translation.